# Creation and Verification of Digital Twins in Cloud Production

*Maksim* Kubrikov[1], *Mikhail* Saramud[1,2], *Angelina* Petetskaya[1] and *Evgeniy* Talay[1]

[1]Reshetnev Siberian State University of Science and Technology 31 Krasnoyarskiy Rabochiy Ave., the City of Krasnoyarsk, the Russian Federation, 660037
[2]Siberian Federal University, 79 Svobodny Ave., the City of Krasnoyarsk, the Russian Federation, 660041

**Abstract**. This article discusses the use of digital twins for products made of polymer composite materials. The design of new products from polymer composite materials, both within the framework of the traditional and new direction of cloud production, requires the need to calculate the physical and mechanical characteristics of the product at the design stage. Carrying out full-scale tests increases greatly the cost and slows down the production. It requires the manufacture of a prototype of the product. The use of existing development tools does not always provide the required characteristics. To solve this problem, it is proposed to use a digital twin, which will not only solve the problem, but will also help to move to cloud production and the development of the *Industry 4.0* direction. Thus, a new problem arises – how to create digital twins of products from polymer composite materials. The analysis makes it possible to conclude that the traditional methods of mathematical physics are not suitable for the solution of this problem, since the twins obtained with their help do not have any properties of adaptability. To solve the problem, it is proposed to use deep neural networks – one of the most powerful methods of machine learning. This will make it possible to obtain digital twins of products made of polymer composite materials that can adapt to changes in the model, environmental conditions and adapt to changes in the indicators of sensors and transducers installed on the product.

## 1 Introduction

The high rates of development of various industries cause an increase in the volume of mass-produced products, which must meet all the requirements and standards presented today. Every year there is an increasing tendency to replace traditional structural materials with modern and promising ones, among which the accent is given to the composite materials.
A polymer composite material (PCM) is a material consisting of two or more components with a pronounced boundary between a binder in the form of a polymer (epoxy resins, polyester resins, etc.) and a reinforcing element (carbon fiber, fiberglass, etc.). Due to the difference in the physical and chemical properties of the components of the composite material, high mechanical characteristics are provided, such as specific strength, stiffness, etc. The combination of these characteristics with a low specific gravity predetermined its widespread use in many areas. In aviation and rocket and space technology, PCMs have become the basis of many designs. For example, spacecraft (SC) reflector antenna mirrors, made of carbon fiber and cyanate-ether binder, make it possible to obtain precise product dimensions, rigidity and dimensional stability.

In recent years, smart composites or self-organizing systems have become very popular. Smart composites can read an external signal, process it and perform an action, and can also have feedback mechanisms, self-diagnostics and self-healing. Smart materials also include flexible composites, shape memory products and other functional materials.

The strength properties of a PCM product depend on many variables, such as binder and matrix materials, the number of layers, reinforcement angles, manufacturing technology (manual laying, vacuum bag, RTM method, etc.), porosity, etc. CAD models of the product are calculated in such CAE packages as ANSYS Composite, Digimat, Siemens NX (ESI Group – Composites Simulation Suite, FiberSim), ABAQUS, etc. Due to the peculiarity of PCM manufacturing, it is not always possible to achieve the designed properties. In this regard, the results of virtual tests on a computer will not correspond to the reality. The use of digital twins (DT) is a promising solution concerning this problem.

## 2 Digital Twin

There are many terms for the digital twin concept. One of the terms states that each object can be represented as a physical and virtual system, so that a virtual model represents a physical one, and vice versa. The concept of the interaction of a physical object in the real world with its digital twin in the virtual space and the presence of direct communication and feedback between them reflect the idea of a DT (Figure 1).

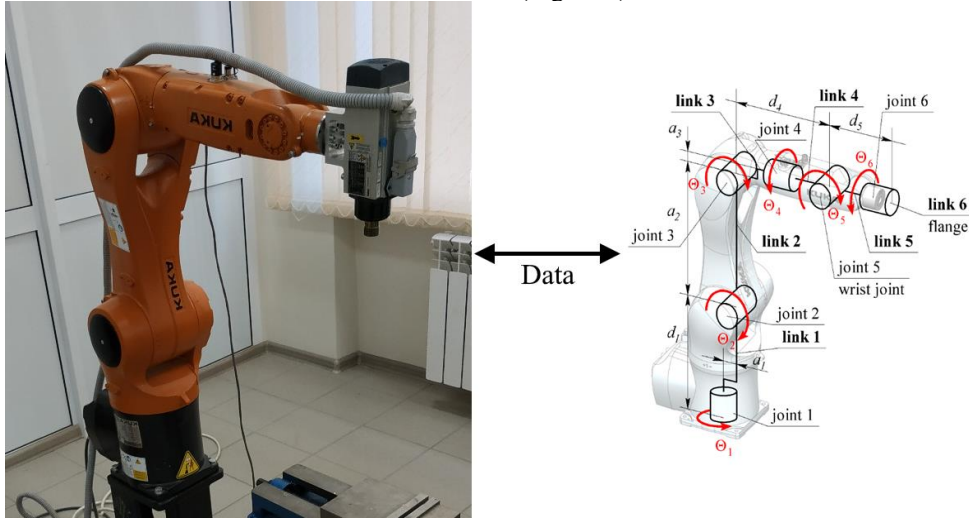

**Fig. 1.** An example of a digital twin

In a review article [1], the authors examined an extensive list of relevant publications. The application of DT in various fields such as health care [2], meteorology [3], production [4], education, transport, energy [5], is considered. It is concluded that the digital twin plays a transformative role not only in how cyber-physical intelligent systems are designed and operated, but also in how the modularity of multidisciplinary systems is advancing to overcome fundamental barriers that cannot be solved by current methods of evolutionary modeling.

Based on the updated model of the PCM DT, recommendations can be formed to optimize the operation and maintenance of a real object. For example, the model can predict the potential for failure of a specific component, recommend times for preventive maintenance, inspections, and other key activities.

In some definitions of the DT, the issues at hand are only about the virtual part, which is defined as a virtual model consisting of two parts – a digital master and a digital shadow.

The digital master must contain comprehensive information sufficient to manufacture a product with certain properties, including an analysis of the physical and mechanical properties of this product. The digital shadow that a physical object "throws on virtual space" is a set of data obtained from sensors and a model that makes it possible to predict the properties of an object within certain limits.

It is assumed that by 2022, about half of the companies will use digital twins of different levels. Simple-level digital twins do not use machine learning, while machine learning algorithms are needed to create a high-level digital twin.

The development of cyber-physical systems, the internet of things, cloud and cognitive computing makes it possible to increase the level of automation, thereby replacing a person in complex and dangerous areas of production. This transition marks the beginning of the fourth industrial revolution (*Industry 4.0*), which leads to digital manufacturing. Information flows between sensors, devices and people complete through the industrial internet of things and people. Digital manufacturing models create low-level data from sensors with high-level contextual information. Based on this information, cyber-physical systems make decisions instantly and autonomously and at the same time provide the human operator with the collected information in a convenient visual form.

The DT of a PCM product reproduces and simulates a structure of several layers of reinforcing material, so that the necessary virtual tests can be carried out in order to find out what physical and mechanical characteristics it has, and thereby reduce the cost of the product by optimizing it. All this is possible through the use of digital twins technology, creating a digital version of the PCM, it is possible to develop a product model and conduct virtual tests.

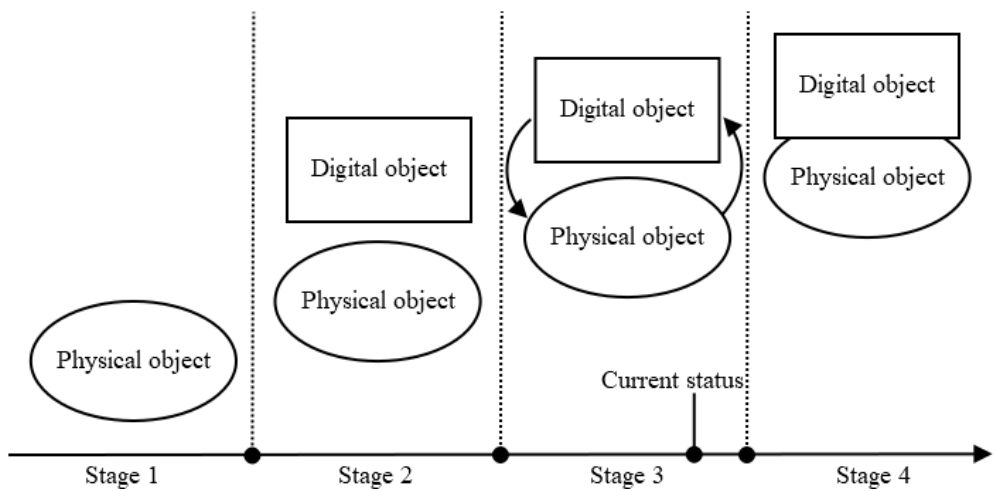

**Fig. 2.** Evolution stages of digital twins

Lior Kitain in article [6] shows the process of evolution of digital twins (Figure 2). In the diagram, the first stage corresponds to the period when physical objects were created without a digital clone. Stage 2 refers to the period when artificial objects were designed using a digital model, which was used only at the stage of object creation. At stage 3, interaction (information exchange) between the physical and digital twins begins. And at stage 4, there is a convergence and so called intersection of the physical and digital twins, when information exchange and updating of the digital and physical twins takes place almost in real time.

## 3 Generation of Digital Twins of PCM Products

Traditionally, methods of mathematical physics are used to create digital models of physical objects. These methods are used in the analysis of the strength characteristics of structures, in heat engineering, hydrodynamics, electric power industry, etc. The problem of modeling a physical object is represented as a set of boundary value problems for ordinary differential equations and (or) differential equations in partial derivatives, for example, for the Euler-Lagrange equations, Navier-Stokes equations.

Unfortunately, these methods do not have a number of important properties necessary for constructing digital twins, such as:
- failure of continuous learning throughout the entire product lifecycle;
- failure of pre-learning and post-learning of the digital model;
- inability to adapt the model as a result of the acquisition of information from sensors;
- failure to predict failures to prevent them;
- impossibility of solving problems of recognition and diagnostics;
- they do not take into account the metrological characteristics of sensors;
- they are unable to automatically change the design of the model, modify and adjust meshes and elements based on observations.

When designing products from PCM, it is important to update models on the go, since new data improve the quality of the final result.

Another approach seems to be more promising, when an adaptive model is built for each element of the target object, which can be refined and rebuilt in accordance with the observational data of the object. It is most convenient to build such a model using a neural network. Therefore, an urgent task is the development of neural network modeling technologies, a more complete accounting of historical and newly received data, improvement of methods for automatic adjustment of the architecture and model parameters, methods of classification and prediction. The set of tasks facing the digital twin model can be solved using a collective of neural networks, each of which displays a certain fragment (element, process) of the object.

Deep neural networks (DNN) are one of the most powerful machine learning techniques. DNNs are inspired by the architecture of biological neural networks that mimic humans learning from acquired data. DNNs usually have an input layer, hidden layers and an output layer. Several researches used DNNs to study the behavior of materials and structures.

In article [7], D. T. Do, D. Lee, J. Lee used deep neural networks to replace finite element analysis in solving optimization problems (bending and free vibration with different volume constraints) of functionally graded plates. In work [8], T. Nguyen, A. Kashani, T. Ngo, S. Bordas used a deep neural network model to predict the compressive strength of foam concrete.

In work [9], A.N. Vasiliev, D.A. Tarkhov and G.F. Malykhina developed methods to build digital twins of real objects. The authors presented algorithms for constructing digital twin models that can be adapted for a specific type of real objects for which a digital twin needs to be built. A special feature of this approach to evolutionary algorithms is the use of genetic procedures for constructing the structure of the model and nonlinear optimization algorithms for adjusting its parameters. The authors also propose an approach to building multilayer models using differential equations, which makes it possible to do without neural network training.

In article [10], Chen, Chun-Teh & Gu, Grace reviewed the application of machine learning to the modeling and design of composite materials. They implicated basic machine learning algorithms in the context of materials design, such as linear regression, neural networks, convolutional neural networks, and Gaussian process.

## 4 Conclusion

It can be concluded that the creation of digital twins of PCM products is an important task that is being solved all over the world. Digital twins will allow predicting the state of the composite and promptly responding to any emergency situations during its operation at the stage of the product life cycle.

The use of digital twins will allow designers at the design stage to accurately determine the characteristics of a future product using a validated base of PCM properties, and operators using a product for its intended purpose will see its state in real time, predict its future state and, if necessary, respond to changes in a timely manner.

The creation of digital twins is of particular interest to institutions engaged in the production and operation of composite products in the field of architecture, rocket and space industry and energy engineering.

This work was supported by the Ministry of Science and Higher Education of the Russian Federation (State Contract No. FEFE-2020-0017).